\documentclass[a4paper,conference]{IEEEtran}
\usepackage{cite}
\usepackage{amsmath,amssymb,amsfonts}
\usepackage{algorithmic}
\usepackage{textcomp}
\usepackage{xcolor}
\usepackage{multirow}
\usepackage{caption}
\usepackage{balance}
\usepackage{array}
\usepackage[final]{graphicx}
\usepackage{subcaption}
\usepackage{pgfplots}
\usepackage{pgfplotstable}
\usepackage{sfmath}
\usepackage{lipsum}
\usepackage{graphicx}
\usepackage{color,colortbl}
\usepackage{booktabs}
\begin{document}
\newpage
\bstctlcite{IEEEexample:BSTcontrol}
\title{Impact of Action Unit Occurrence Patterns on Detection}

\author{\IEEEauthorblockN{Saurabh Hinduja}
\IEEEauthorblockA{Computer Science and Engineering\\
University of South Florida\\
Tampa, Florida 33620\\
Email: saurabhh@mail.usf.edu}
\and
\IEEEauthorblockN{Shaun Canavan}
\IEEEauthorblockA{Computer Science and Engineering\\
University of South Florida\\
Tampa, Florida 33620\\
Email: scanavan@usf.edu}
\and\IEEEauthorblockN{Saandeep Aathreya}
\IEEEauthorblockA{Computer Science and Engineering\\
University of South Florida\\
Tampa, Florida 33620\\
Email: saandeepaath@mail.usf.edu}}
\maketitle

\begin{abstract}
Detecting action units is an important task in face analysis, especially in facial expression recognition. This is due, in part, to the idea that expressions can be decomposed into multiple action units. In this paper we investigate the impact of action unit occurrence patterns on detection of action units. To facilitate this investigation, we review state of the art literature, for AU detection, on 2 state-of-the-art face databases that are commonly used for this task, namely DISFA, and BP4D. Our findings, from this literature review, suggest that action unit occurrence patterns strongly impact evaluation metrics (e.g. F1-binary). Along with the literature review, we also conduct multi and single action unit detection, as well as propose a new approach to explicitly train deep neural networks using the occurrence patterns to boost the accuracy of action unit detection. These experiments validate that action unit patterns directly impact the evaluation metrics. 

\end{abstract}
\section{Introduction}
Facial expression recognition is a growing field that has attracted the attention of many research groups due in part from early work from Picard \cite{picard2000affective}. A range of modalities have been found useful for this problem including 2D \cite{yang2018facial, chen2018facial}, thermal \cite{wang2018thermal, nguyen2018towards} and 3D/4D data \cite{chang2018expnet, cheng20184dfab}. Promising multimodal approaches have also been proposed \cite{li2018mec, wei2018unsupervised}. Another interesting approach is based on the detection of action units (AU). These works are based on the Facial Action Coding System \cite{friesen1978facial} (FACS), which is seminal work that decomposes facial expressions into a set of action units.
\definecolor{mycolor}{RGB}{255,51,76}
\definecolor{mycolor5}{RGB}{125,120,115}
\definecolor{mycolor15}{RGB}{200,50,115}
\begin{center}
\centering
\pgfplotstableread[col sep=&, header=true]{
AU	&BP4D	&LSTM	&LSVM	    &DAM	&MDA	&GFK	&iCPM	&JPML	&DRML	&FVGG	&E-Net	&EAC	&ROI	&R-T1	&R-T2 &TEMPT-Net    &Ones &D-PattNett &DSIN &JAA-Net
AU 1	&0.21	&0.314	&0.46	&0.382	&0.396	&0.424	&0.466	&0.326	&0.364	&0.278	&0.376	&0.39	&0.36	&0.47	&0.46   &0.42   &0.35   &0.51   &0.51   &0.47
AU 2	&0.18	&0.311	&0.385	&0.273	&0.37	&0.358	&0.387	&0.256	&0.418	&0.276	&0.321	&0.352	&0.32	&0.56	&0.48   &0.3    &0.29   &0.43   &0.51   &0.44
AU 4	&0.21	&0.714	&0.485	&0.291	&0.457	&0.473	&0.465	&0.374	&0.43	&0.183	&0.442	&0.486	&0.43	&0.52	&0.46   &0.32   &0.34   &0.59   &0.58   &0.55
AU 6	&0.47	&0.633	&0.67	&0.675	&0.692	&0.712	&0.684	&0.423	&0.55	&0.697	&0.756	&0.761	&0.77	&0.79	&0.77   &0.72   &0.63   &0.79   &0.77   &0.77
AU 7	&0.56	&0.771	&0.722	&0.726	&0.702	&0.725	&0.738	&0.505	&0.67	&0.691	&0.745	&0.729	&0.74	&0.81	&0.8    &0.72   &0.71   &0.79   &0.74   &0.75
AU 10	&0.62	&0.45	&0.7277	&0.744	&0.71	&0.742	&0.741	&0.722	&0.663	&0.781	&0.808	&0.819	&0.85	&0.88	&0.85   &0.81   &0.75   &0.85   &0.85   &0.84
AU 12	&0.59	&0.826	&0.836	&0.764	&0.818	&0.839	&0.846	&0.741	&0.658	&0.632	&0.851	&0.862	&0.87	&0.89	&0.87   &0.83   &0.72   &0.83   &0.87   &0.57
AU 14	&0.5	&0.729	&0.599	&0.599	&0.578	&0.572	&0.622	&0.657	&0.541	&0.364	&0.568	&0.588	&0.63	&0.75	&0.72   &0.62   &0.64   &0.67   &0.72   &0.62
AU 15	&0.2	&0.34	&0.411	&0.159	&0.414	&0.406	&0.443	&0.381	&0.332	&0.261	&0.316	&0.375	&0.46	&0.59	&0.48   &0.33   &0.29   &0.51   &0.40   &0.43
AU 17	&0.38	&0.539	&0.556	&0.529	&0.501	&0.554	&0.575	&0.4	&0.48	&0.507	&0.556	&0.591	&0.58	&0.68	&0.6    &0.42   &0.51   &0.65   &0.67   &0.6
AU 23	&0.22	&0.386	&0.408	&0.039	&0.362	&0.399	&0.417	&0.304	&0.317	&0.228	&0.219	&0.359	&0.38	&0.4	&0.38   &0.38   &0.28   &0.49   &0.38   &0.43
AU 24	&0.2	&0.37	&0.421	&0.049	&0.411	&0.417	&0.397	&0.423	&0.3	&0.359	&0.291	&0.358	&0.37	&0.59	&0.51   &0.12   &0.26   &0.57   &0.47   &0.42
}\mydata
\centering
\pgfplotstableread[col sep=&, header=true]{
AU	&DISFA	&LSVM	&APL	&DRML	&FVGG	&E-Net	&EAC	&ROI	&R-T1   &DSIN   &JAA-Net 
AU 1	&0.067	&0.11	&0.11	&0.173	&0.33	&0.37	&0.42	&0.42	&0.43   &0.47   &0.44
AU 2	&0.056	&0.1	&0.12	&0.18	&0.24	&0.06	&0.26	&0.26	&0.27   &0.42   &0.46
AU 4	&0.199	&0.22	&0.3	&0.37	&0.61	&0.47	&0.66	&0.66	&0.66   &0.69   &0.56
AU 6	&0.149	&0.16	&0.12	&0.29	&0.34	&0.53	&0.51	&0.51	&0.56   &0.32   &0.41
AU 9	&0.0545	&0.12	&0.1	&0.11	&0.017	&0.13	&0.81	&0.085	&0.23   &0.52   &0.45
AU 12	&0.235	&0.7	&0.66	&0.38	&0.72	&0.71	&0.89	&0.89	&0.83   &0.73   &0.69
AU 25	&0.352	&0.12	&0.21	&0.39	&0.87	&0.84	&0.89	&0.89	&0.88   &0.92   &0.88
AU 26	&0.19	&0.2	&0.27	&0.2	&0.07	&0.44	&0.16	&0.16	&0.26   &0.46   &0.58
}\dataDISFA
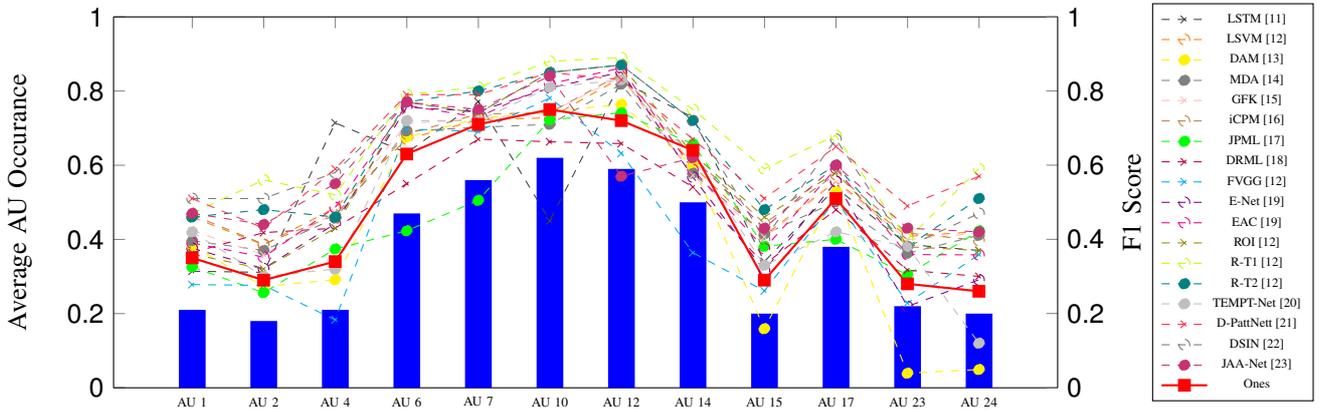
\begin {figure*}
\label{fig:occVsScoreBP4D}
\centering
\begin{tikzpicture}
  \centering
    \begin{axis}
      [
      axis y line*=left,
      ylabel=Average AU Occurance,
        xtick=data,
        xticklabels from table={\mydata}{AU},      
        height=6.5cm, width=14cm,
        ymin=0, ymax=1,
        nodes near coords align={vertical},
                    xticklabel style={font=\tiny} ,
      ]
\addplot  [draw=none, fill=blue] [ybar] table [y=BP4D, x expr=\coordindex,] {\mydata};
\end{axis}
    \begin{axis}
      [
      legend style={font=\fontsize{6}{5}\selectfont},
axis y line*=right,
  axis x line=none,
  ymin=0, ymax=1,
        xtick=data,
        xticklabels from table={\mydata}{AU},      
        height=6.5cm, width=14cm,
        nodes near coords align={vertical},
         ylabel=F1 Score,
         ylabel near ticks, 
         yticklabel pos=right,
         legend style={nodes={scale=0.8, transform shape}, at={(1.1,0.5)},anchor=west}
      ]
\addplot [mark=x,dashed, darkgray] table [y=LSTM, x expr=\coordindex] {\mydata};
\addplot [mark=o,dashed,orange] table [y=LSVM, x expr=\coordindex] {\mydata};
\addplot [mark=*,dashed,yellow] table [y=DAM, x expr=\coordindex] {\mydata};
\addplot [mark=*,dashed,gray] table [y=MDA, x expr=\coordindex] {\mydata};
\addplot [mark=x,dashed,pink] table [y=GFK, x expr=\coordindex] {\mydata};
\addplot [mark=o,dashed,brown] table [y=iCPM, x expr=\coordindex] {\mydata};
\addplot [mark=*,dashed,green] table [y=JPML, x expr=\coordindex] {\mydata};
\addplot [mark=x,dashed,purple] table [y=DRML, x expr=\coordindex] {\mydata};
\addplot [mark=x,dashed,cyan] table [y=FVGG, x expr=\coordindex] {\mydata};
\addplot [mark=o,dashed,violet] table [y=E-Net, x expr=\coordindex] {\mydata};
\addplot [mark=o,dashed,magenta] table [y=EAC, x expr=\coordindex] {\mydata};
\addplot [mark=x,dashed,olive] table [y=ROI, x expr=\coordindex] {\mydata};
\addplot [mark=o,dashed,lime] table [y=R-T1, x expr=\coordindex] {\mydata};
\addplot [mark=*,dashed,teal] table [y=R-T2, x expr=\coordindex] {\mydata};
\addplot [mark=*,dashed,lightgray] table [y=TEMPT-Net, x expr=\coordindex] {\mydata};
\addplot [mark=x ,dashed, mycolor] table [y=D-PattNett, x expr=\coordindex] {\mydata};
\addplot [mark=o ,dashed, mycolor5] table [y=DSIN, x expr=\coordindex] {\mydata};
\addplot [mark=* ,dashed, mycolor15] table [y=JAA-Net, x expr=\coordindex] {\mydata};
\addplot [mark=square*, thick, red] table [y=Ones, x expr=\coordindex] {\mydata};

\addlegendentry{LSTM\cite{chu2017learning}}
\addlegendentry{LSVM\cite{li2017action}}
\addlegendentry{DAM\cite{duan2009domain}}
\addlegendentry{MDA\cite{sun2011two}}
\addlegendentry{GFK\cite{gong2012geodesic}}
\addlegendentry{iCPM\cite{zeng2015confidence}}
\addlegendentry{JPML\cite{zhao2015joint}}
\addlegendentry{DRML\cite{zhao2016deep}}
\addlegendentry{FVGG\cite{li2017action}}
\addlegendentry{E-Net\cite{li2017eac}}
\addlegendentry{EAC\cite{li2017eac}}
\addlegendentry{ROI\cite{li2017action}}
\addlegendentry{R-T1\cite{li2017action}}
\addlegendentry{R-T2\cite{li2017action}}
\addlegendentry{TEMPT-Net\cite{liu2019multi}}
\addlegendentry{D-PattNett\cite{onal2019d}}
\addlegendentry{DSIN\cite{corneanu2018deep}}
\addlegendentry{JAA-Net\cite{shao2018deep}}
\addlegendentry{Ones}
\end{axis}
\end{tikzpicture}
    \caption{AU detection accuracy vs occurrence in BP4D. Bars are the average number of AU occurrences, per frame, across all subjects. Line graphs are different F1 scores, of methods in the literature, for each AU. NOTE: "Ones" is what happens when we manually label all 1's (i.e. AU is active) for each of the AUs. \emph{NOTE: Best viewed in color}.}
  \label{graph:DistributionBP4D}
 \vspace{-2mm}
\end{figure*}
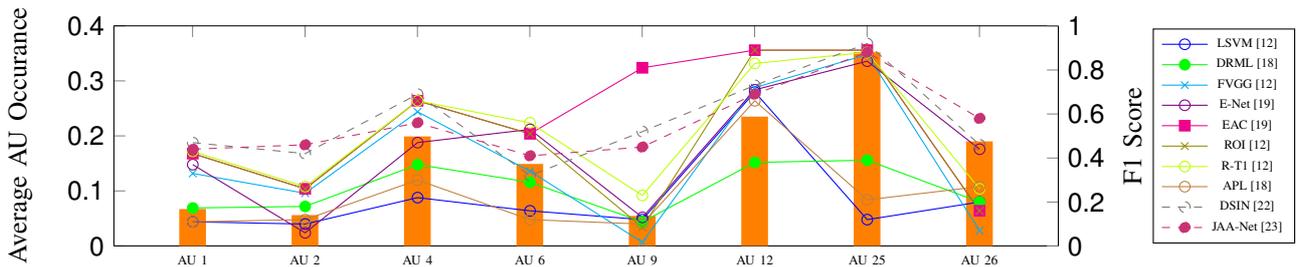
\begin {figure*}
\centering
\begin{tikzpicture}
  \centering
    \begin{axis}
      [
      axis y line*=left,
      ylabel=Average AU Occurance,
        xtick=data,
        xticklabels from table={\dataDISFA}{AU},      
        height=4.5cm, width=14cm,
        ymin=0, ymax=0.4,
        nodes near coords align={vertical},
                    xticklabel style={font=\tiny} ,
      ]
\addplot  [draw=none, fill=orange] [ybar] table [y=DISFA, x expr=\coordindex,] {\dataDISFA};
\end{axis}
    \begin{axis}
      [
      legend style={font=\fontsize{6}{5}\selectfont},
axis y line*=right,
  axis x line=none,
  ymin=0, ymax=1,
        xtick=data,
        xticklabels from table={\dataDISFA}{AU},      
        height=4.5cm, width=14cm,
        nodes near coords align={vertical},
         ylabel=F1 Score,
         ylabel near ticks, 
         yticklabel pos=right,
         legend style={nodes={scale=0.8, transform shape}, at={(1.1,0.5)},anchor=west}
      ]
\addplot [mark=o,blue] table [y=LSVM, x expr=\coordindex] {\dataDISFA};
\addplot [mark=*,green] table [y=DRML, x expr=\coordindex] {\dataDISFA};
\addplot [mark=x,cyan] table [y=FVGG, x expr=\coordindex] {\dataDISFA};
\addplot [mark=o,violet] table [y=E-Net, x expr=\coordindex] {\dataDISFA};
\addplot [mark=square*, magenta] table [y=EAC, x expr=\coordindex] {\dataDISFA};
\addplot [mark=x, olive] table [y=ROI, x expr=\coordindex] {\dataDISFA};
\addplot [mark=o, lime] table [y=R-T1, x expr=\coordindex] {\dataDISFA};
\addplot [mark=o,brown] table [y=APL, x expr=\coordindex] {\dataDISFA};
\addplot [mark=o ,dashed, mycolor5] table [y=DSIN, x expr=\coordindex] {\dataDISFA};
\addplot [mark=* ,dashed, mycolor15] table [y=JAA-Net, x expr=\coordindex] {\dataDISFA};
\addlegendentry{LSVM\cite{li2017action}}
\addlegendentry{DRML\cite{zhao2016deep}}
\addlegendentry{FVGG\cite{li2017action}}
\addlegendentry{E-Net\cite{li2017eac}}
\addlegendentry{EAC\cite{li2017eac}}
\addlegendentry{ROI\cite{li2017action}}
\addlegendentry{R-T1\cite{li2017action}}
\addlegendentry{APL\cite{zhao2016deep}}
\addlegendentry{DSIN\cite{corneanu2018deep}}
\addlegendentry{JAA-Net\cite{shao2018deep}}
\end{axis}
\end{tikzpicture}
    \caption{AU detection accuracy vs occurrence in DISFA. Bars are the average number of AU occurrences, per frame, across all subjects. Line graphs are different F1 scores, of methods in the literature, for each AU. NOTE: The scales on the left and right are different due to the low number of active AUs compared to some of the F1 scores. \emph{NOTE: Best viewed in color}.}
  \label{graph:DistributionDISFA}
  \vspace{-2mm}
\end{figure*}
\end{center}

 \vspace{-\baselineskip}
  \vspace{-\baselineskip}
There have been promising approaches to AU detection that have made use of both hand-crafted features, as well as deep learning. Liu et al. \cite{liu2019multi} proposed TEMT-NET that uses the correlations between AU detection and facial landmarks. Their proposed network performs action unit detection, facial landmark detection, and thermal image reconstruction simultaneously. The thermal reconstruction and facial landmark detection provide regularization on the learned features providing a boost to AU detection performance. Werner et al. \cite{werner2017facial} investigated action unit intensity estimation using modified random regression forests. They also developed a visualization technique for the relevance of the features. Their results suggest that precomputed features are not enough to detect certain AUs. Li et al. \cite{li2018eac} proposed the EAC-Net, which is a deep network that enhances and crops regions of interest for AU detection. They found the proposed approach allows for robustness to face position and pose. They also integrate facial attention maps that correspond to areas of the face with active AUs. Their results suggest using these attention maps can enhance the learning in the network, at specific layers. 
The Facial Expression Recognition and Analysis challenge (FERA) \cite{valstar2011first, ValstarFera15, valstFERA17} also focused on the detection of AUs. This challenge was designed to address the challenge of a common evaluation protocol for AU detection.

Girard et al. \cite{girard2015much} looked at how much training data is needed for AU detection. They investigated 80 subjects, and more than 350,000 frames of data using SIFT features. Their results suggest that variation in the total number of subjects is an important factor in increasing AU detection accuracies, as they achieved their max accuracy from a large number of subjects. Ertugrul et al. \cite{ertugrulFG2019} investigated the efficacy of cross-domain AU detection. Using shallow and deep approaches, their results suggest that more varied domains, as well as deep learning can increase generalizability. Jeni et al. \cite{jeni2013facing} evaluated the effect of imbalanced data on AU detection. They evaluated accuracy, F-score, Cohen's kappa, Krippendorf's alpha, receiver operator characteristic (ROC) curve, and precision-recall curve. Their findings indicate that the skewed data attenuated all metrics except for ROC, however, it may mask poor performance and they recommend reporting skew-normalized scores.

Although these works have shown success in detecting AUs, to the best of our knowledge, none have investigated the impact of AU occurrence patterns on the accuracy of detection. Motivated by this, we specifically focus our experiments on AU occurrence patterns, showing they have significant impact on accuracy. Considering this, the contribution of our work is 3-fold, and can be summarized as follows:
\begin{enumerate}
    \item In-depth analysis on the impact of AU occurrence patterns on detection results, showing the patterns directly impact evaluation metrics, especially F1-binary. 
    \item Multi and single AU detection experiments are conducted validating that AU occurrence patterns directly impact evaluation metrics. We have shown that a network that performs poorly on single AU detection, can learn the patterns thus improving overall performance.
    \item A new method for AU detection is proposed that directly trains deep neural networks on occurrence patterns to help mitigate their negative impact on AU detection accuracy. 
\end{enumerate}
\section{Action Unit Occurrence Patterns}
\label{sec:distribution}

\subsection{Datasets}
\label{sec:data}
\textbf{DISFA} is a spontaneous dataset for studying facial action unit intensity. It contains 27 adults (12 women/15 men) that watched a 4-minute video clip meant to elicit spontaneous expressions. For our analysis, all frames from this dataset are used, as all are AU annotated (130,815 frames). 

\textbf{BP4D} is a multimodal (e.g. 2D, 3D, AUs) facial expression dataset which was used in the Facial Expression Recognition and Analysis challenges in 2015 \cite{ValstarFera15} and 2017 \cite{valstFERA17} where the focus was the detection of occurrence and intensity of AUs. There are a total of 41 subjects (23 female/18 male) displaying eight dynamic expressions (happy, surprise, sad, startled, skeptical, embarrassed, fear, and pain). For our analysis, all AU annotated frames are used from this dataset (146,847 frames). 

\subsection{Class Imbalance and AU Detection}
\label{sec:imbalAURec}

It has been shown that AU class imbalance has a negative impact on evaluation metrics \cite{jeni2013facing}. To further extend this notion, we will show that class imbalance and AU occurrence patterns are directly correlated. In order to analyze the impact of occurrence patterns on AU detection, we must first analyze the imbalance of AU classes. To facilitate this, we have analyzed state-of-the-art literature regarding their F1-binary scores from AU detection experiments on BP4D and DISFA. As can be seen in Figs. \ref{graph:DistributionBP4D} and \ref{graph:DistributionDISFA}, the occurrence of AUs in both datasets are imbalanced. For example, in BP4D, AU 10 has an average occurrence of 0.62, while AU 2 has an average AU occurrence of 0.18. There is a direct correlation between these occurrences and their F1-binary scores. AU 10 is one of the highest occurring AUs and also one of the highest F1-binary scores across the literature, with an average F1-binary score of 0.75. Similarly, AU 2 is one of the lowest occurring AU and one of the lowest F1-binary scores across the literature, with an average F1-binary score of 0.36. Considering this, our analysis shows that current state-of-the-art results, in AU detection, follow an explicit trend which is correlated to the imbalance of the AUs. To further illustrate this trend, we also calculated the F1-binary score if we were to manually predict all 1's, for all frames (i.e. all AUs are active). As can be seen in Fig. \ref{graph:DistributionBP4D}, the general trend that the F1-binary scores follow, for all methods in BP4D, is the same as labeling all 1's.
\begin{table}
\newcolumntype{P}[1]{>{\centering\arraybackslash}p{#1}}
\centering
\captionsetup{justification=centering}
\caption{Correlation between F1-binary scores and AU class imbalance.}
{\fontsize{6.5}{7}\selectfont
\begin{tabular}{ |P{2cm}||P{1cm}|P{1cm}|  }
 \hline 
 \multirow{2}{*}{Method} & \multicolumn{2}{c|}{Correlation}\\ \cline{2-3}&BP4D &DISFA  \\ \hline 						

LSTM\cite{chu2017learning}	&0.680	&-\\ \hline 
LSVM\cite{li2017action}	&0.957	&0.347	\\ \hline 
DAM\cite{duan2009domain}	&0.922	&-\\ \hline 
MDA\cite{sun2011two}	&0.948	&-\\ \hline 
GFK\cite{gong2012geodesic}	&0.951	&-\\ \hline 
iCPM\cite{zeng2015confidence}	&0.967	&-\\ \hline 
JPML\cite{zhao2015joint}	&0.869	&-\\ \hline 
DRML\cite{zhao2016deep}	&0.949	&0.844	\\ \hline 
FVGG\cite{li2017action}	&0.890	&0.785	\\ \hline 
E-Net\cite{li2017eac}	&0.944	&0.919	\\ \hline 
EAC\cite{li2017eac}	&0.953		&0.472	\\ \hline 
ROI\cite{li2017action}	&0.966		&0.773	\\ \hline 
R-T1\cite{li2017action}	&0.931	&0.816\\ \hline 
R-T2\cite{li2017action}	&0.970	&-\\ \hline 
APL\cite{zhao2016deep}	&- &0.5098\\ \hline 
D-PattNett\cite{onal2019d} &0.946 &-\\ \hline
DSIN\cite{corneanu2018deep} &0.931 &0.792 \\ \hline
JAA-Net\cite{shao2018deep} &0.847 &0.918 \\ \hline
\textbf{Average} &\textbf{0.916}	&\textbf{0.718}\\ \hline 
\end{tabular}
}
\vspace{-2.5mm}
\label{table:MethodDataCorrelation}
\end{table}
The general trend that F1-binary scores follow can visually be seen in  Figs. \ref{graph:DistributionBP4D} and \ref{graph:DistributionDISFA}. To statistically analyze this trend, we calculated the correlation between the class imbalance and F1-binary scores of the methods shown in Figs. \ref{graph:DistributionBP4D} and \ref{graph:DistributionDISFA}. We define correlation as $corr=\dfrac{\sum{(x-\overline{x})(y-\overline{y})}}{\sqrt{\sum{{(x-\overline{x})^2}{(y-\overline{y})^2}}}}$ , where $\overline{x}$ and $\overline{y}$ are the averages of the classes and the F1-binary scores, respectively. For BP4D and DISFA, the average correlations are 0.921 and 0.683, respectively. These results suggest that there is high correlation between the class imbalance and the reported F1-binary scores for AU detection (Table \ref{table:MethodDataCorrelation}). This further suggests that AUs with high F1-binary scores have a lot of training data, and AUs with low F1-binary scores have a small amount. Although there is a general trend of AU occurrence versus F1-binary accuracy, it is important to note there are some anomalies in the F1-binary scores for some AUs and methods. For example, on BP4D, Chu et al \cite{chu2017learning} use high intensity AUs, equal to or higher than A-level for positive samples, and the rest for negative. In DISFA, some of the experiments train on BP4D and test on DISFA, which is a common approach for testing on this dataset, due to the imbalance of AUs. This can explain, in part, some of the lower correlations in Table \ref{table:MethodDataCorrelation} (e.g. \cite{chu2017learning, li2017eac, li2017action}).

Along with the correlation between the class imbalance and F1-binary scores, we also calculated the variance, $var=\dfrac{\sum{{(x-\overline{x})}^2}}{(n-1)}$, and standard deviation, $std=\sqrt{var}$, of the F1-binary scores between each of the methods detailed in Figs. \ref{graph:DistributionBP4D} and \ref{graph:DistributionDISFA}. There is a small amount of variance between each of the methods across all studied AUs. In BP4D, the average variance is 0.0089 (std of 0.0923), and in DISFA the average variance is 0.0383 (std of 0.180). This suggests that the investigated F1-binary scores are all similar within a small accuracy range. While the general variance and standard deviations are low, there are some outliers, especially in DISFA. For example, AU 9 has a variance of 0.064 and standard deviation of 0.253. This can also be visually seen in Fig. \ref{graph:DistributionDISFA}, with the F1-binary score of Li et al. \cite{li2017eac}. As previously detailed, this can partially be explained from training on BP4D and testing on DISFA. See Table \ref{table:VArianceBetweenMethods} for the standard deviation between all methods. This analysis naturally leads to the question of why is this specific trend occurring in AU detection. We hypothesized that this is due to the AU occurrence patterns that occur across all of the sequences in the data.

\begin{table}
\centering
\captionsetup{justification=centering}
\caption{Standard deviation of F1-binary scores, for each individual AU, between all methods (Figs. \ref{graph:DistributionBP4D}, and \ref{graph:DistributionDISFA}).}
\label{table:VArianceBetweenMethods}
\newcolumntype{P}[1]{>{\centering\arraybackslash}p{#1}}

{\fontsize{6.5}{8}\selectfont
\begin{tabular}{|c|c|c|}
\hline
\multirow{2}[4]{*}{AU} & \multicolumn{2}{c|}{Std Dev} \\
\cline{2-3}      & BP4D  & DISFA \\
\hline
AU 1  & 0.0671 & 0.1418 \\
\hline
AU 2  & 0.0853 & 0.1301 \\
\hline
AU 4  & 0.1197 & 0.1702 \\
\hline
AU 6  & 0.0935 & 0.1559 \\
\hline
AU 7  & 0.0660 & - \\
\hline
AU 9  & -     & 0.2541 \\
\hline
AU 10 & 0.1006 & - \\
\hline
AU 12 & 0.0917 & 0.1455 \\
\hline
AU 14 & 0.0885 & - \\
\hline
AU 15 & 0.0965 & - \\
\hline
AU 17 & 0.0847 & - \\
\hline
AU 23 & 0.0663 & - \\
\hline
AU 24 & 0.1076 & - \\
\hline
AU 25 & -     & 0.3172 \\
\hline
AU 26 & -     & 0.1573 \\
\hline
\textbf{Average} & \textbf{0.0890} & \textbf{0.1840} \\
\hline
\end{tabular}%
}
\vspace{-3mm}
\end{table}

\subsection{Action Unit Occurrence Patterns}
\begin{figure*}

\pgfplotsset{width=4cm,compat=1.8}
\renewcommand*{\familydefault}{\sfdefault}
\begin{tikzpicture}
  \centering

  \begin{axis}[
   ybar,
    ymin=0, ymax=1,
        xtick=data,
        nodes near coords align={vertical},
         ylabel=Score,
         ylabel near ticks, 
         yticklabel pos=left,
        height=4cm, width=17cm,
        bar width=0.4cm,
        ymin=0, ymax=40,
        legend style={
            at={(0.14,0.95)},
            anchor=north,
            legend columns=-1,
            /tikz/every even column/.append style={column sep=0.5cm}
        },
        xticklabel style={font=\tiny} ,
        yticklabel style={font=\tiny} ,
        ylabel={Percentage of patterns},
        xlabel={Frame Count Range},
        symbolic x coords={5000-11000,2000-5000,1000-2000,500-1000,200-500,100-200,50-100,10-50,5-10,0-5 },
       xtick=data,
       nodes near coords={
       }
    ]
    \addplot [draw=none, fill=blue] coordinates {
      (5000-11000,0.11820331)
      (2000-5000,0.059101655) 
      (1000-2000,1.182033097)
      (500-1000,1.477541371) 
      (200-500,5.082742317) 
      (100-200,7.624113475)
      (50-100,11.5248227) 
      (10-50,33.27423168)
      (5-10,13.4751773)
      (0-5,26.1820331)};
   \addplot [draw=none,fill=orange] coordinates {
      (5000-11000,1.515151515)
      (2000-5000,0.757575758) 
      (1000-2000,2.272727273)
      (500-1000,3.787878788) 
      (200-500,7.196969697) 
      (100-200,7.954545455)
      (50-100,10.60606061) 
      (10-50,30.3030303)
      (5-10,13.25757576)
      (0-5,22.34848485)};
      
;

    \legend{BP4D,DISFA}
  \end{axis}

  \end{tikzpicture}
  \vspace{-2mm}
    \caption{Patterns and their frame counts, compared to total percentage of patterns for BP4D and DISFA. }
  \label{graph:patternFreq}
  \vspace{-5mm}
\end{figure*}
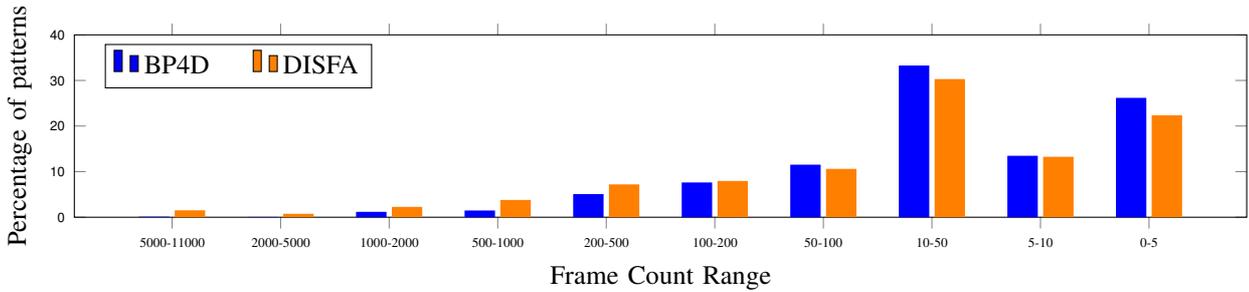
In Section \ref{sec:imbalAURec}, we analyzed the common AUs found in the literature. Here, the AUs investigated are the same for BP4D (see Table \ref{table:VArianceBetweenMethods}), however, for DISFA we are investigating all available AUs (1, 2, 4, 5, 6, 9, 12, 15, 17, 20, 25, 26). We are interested in which AU patterns occur most often, where a pattern is defined as the vector $v = [AU_1, \dots, AU_k]$, where each element can have the values $\{0,1\}$, and k is the number of AUs of interest (e.g. $k=12$ for DISFA). To investigate this, we looked at how many individual patterns exist, as well as the total frame count of each pattern (i.e. how many frames of data have that pattern). There are 1692 different patterns in BP4D and 457 in DISFA, that contain the investigated AUs for each dataset respectively. Tables \ref{table:patternBP4D} and \ref{table:patternDISFA} show the 5 patterns with the highest, and the 2 with the lowest frame counts in BP4D and DISFA, respectively.

Tables \ref{table:patternBP4D} and \ref{table:patternDISFA} reveal a significant imbalance in not only the number of individual AUs, but also the total number of AU occurrence patterns. As can be seen in Table \ref{table:patternBP4D}, the top pattern in BP4D is all 0's. There are 10,630 frames ($\approx 7.2\%$ of the AU annotated frames) where there are no active AUs. Similar, all 0's is the top pattern in DISFA, as can be seen in Table \ref{table:patternDISFA}. Here, there are 48,616 frames that have no active AUs ($\approx 37\%$ of the data). As can be seen in Table \ref{table:patternBP4D}, the bottom 2 patterns, of BP4D, both occur only once, while the top pattern, that has active AUs, occurs 8402 times. Similarly, in DISFA (Table \ref{table:patternDISFA}) the bottom 2 patterns only occur once, while the top pattern with active AUs occurs 8305 times. While there are less overall patterns in DISFA, there is still a large imbalance in the active and inactive AUs per frame. This difference in the number of patterns contributes, at least partially, to the imbalance of AUs in these datasets. It is also important to note that for both datasets, there is a similar trend in the percentage of patterns that exist in a range of frame counts (Fig. \ref{graph:patternFreq}). For example, $\textgreater 72\%$ of the patterns in BP4D and $\textgreater 65\%$ of the patterns in DISFA occur $\textless 50$ times in both datasets. On the other hand, $\textless 0.2\%$ of the patterns in BP4D and $\textless 2\%$ of the patterns in DISFA occur $\textgreater 5000$ times. Our analysis suggests, this imbalance of patterns has a direct impact on the evaluation metrics as there are many patterns that take up a large percentage of the total patterns, however, each of these patterns has a small number of instances that can effectively be used for training machine learning classifiers.
\begin{table}
\centering
\captionsetup{justification=centering}
\caption{BP4D AU occurrence patterns and total number of frames for each pattern. \textit{NOTE: top 5 and bottom 2 rows are patterns with highest and lowest counts, respectively.}}
\newcolumntype{P}[1]{>{\centering\arraybackslash}p{#1}}
{\fontsize{6.5}{7.5}\selectfont
\begin{tabular}{|P{1.4cm}||P{0.14cm}|P{0.14cm}|P{0.14cm}|P{0.14cm}|P{0.14cm}|P{0.14cm}|P{0.14cm}|P{0.14cm}|P{0.14cm}|P{0.14cm}|P{0.14cm}|P{0.14cm}|  }
 \hline 
 \multirow{2}{*}{Frame Count} &\multicolumn{12}{c|}{Pattern}\\ \cline{2-13} &1 &2 &4 &6 &7 &10 &12 &14 &15 &17 &23 &24 \\ \hline
  \rowcolor{lightgray}

10630	&0 &0 &0 &0 &0 &0 &0 &0 &0 &0 &0 &0\\ \hline 

8402	&0 &0 &0 &1 &1 &1 &1 &1 &0 &0 &0 &0\\ \hline 
 \rowcolor{lightgray}

6883	&0 &0 &0 &1 &1 &1 &1 &0 &0 &0 &0 &0\\ \hline 

3571	&0 &0 &1 &0 &0 &0 &0 &0 &0 &0 &0 &0\\ \hline 
 \rowcolor{lightgray}

1814	&0 &0 &0 &0 &0 &1 &1 &0 &0 &0 &0 &0\\ \hline 

1	&0 &0 &1 &0 &1 &1 &0 &0 &1 &1 &1 &0\\ \hline
 \rowcolor{lightgray}

1	&1 &1 &0 &1 &1 &1 &1 &0 &1 &1 &1 &1\\ \hline 
 \end{tabular}
 }
\label{table:patternBP4D}
\end{table}
\begin{table}
\centering
\captionsetup{justification=centering}
\caption{DISFA AU occurrence patterns and total number of frames for each pattern. \textit{NOTE: top 5 and bottom 2 rows are patterns with highest and lowest counts, respectively.}}
\newcolumntype{P}[1]{>{\centering\arraybackslash}p{#1}}
{\fontsize{6.5}{7}\selectfont
\begin{tabular}{ |P{1.4cm}||P{0.14cm}|P{0.14cm}|P{0.14cm}|P{0.14cm}|P{0.14cm}|P{0.14cm}|P{0.14cm}|P{0.14cm}| P{0.14cm}|P{0.14cm}|P{0.14cm}|P{0.14cm}| }
 \hline 
 \multirow{2}{*}{Frame Count} &\multicolumn{12}{c|}{Pattern}\\ \cline{2-13} &1 &2 &4 &5 &6 &9 &12 &15 &17 &20 &25 &26\\ \hline
  \rowcolor{lightgray}

48616	&0	&0	&0	&0	&0	&0		&0	&0  &0	&0  &0	&0\\ \hline
8305	&0	&0	&0	&0	&0	&0		&0	&0  &0  &0  &1  &0\\ \hline
 \rowcolor{lightgray}

6035	&0	&0	&1	&0	&0	&0		&0	&0  &0	&0  &0	&0\\ \hline
5167	&0	&0	&0	&0	&1	&0		&1	&0  &0	&0  &1	&0\\ \hline
 \rowcolor{lightgray}

4994	&0	&0	&0	&0	&0	&0		&1	&0  &0	&0  &1	&1 \\ \hline
1	    &1	&1	&1	&0	&0	&0		&0	&1  &1	&1  &0	&0\\ \hline
 \rowcolor{lightgray}

1	    &0	&1	&0	&1	&0	&0		&0	&0  &0	&1  &1	&1\\ \hline
 \end{tabular}
 }
\label{table:patternDISFA}
\vspace{-2mm}
\end{table}

We also analyzed the AU occurrence patterns as they relate to the sequences and tasks in DISFA and BP4D. Each of the sequences/tasks corresponds to a specific emotion that was meant to be elicited. For each sequence or task, we hypothesized that different, specific patterns would exist for each of them based on FACS \cite{FACS}. Our analysis, of these patterns on BP4D and DISFA, does not support this hypothesis, as can be seen in Tables \ref{table:seqPatternDISFA} and \ref{table:seqPatternBP4D}. In BP4D, it can be seen that the patterns with the highest frame count for tasks T2 and T3 (surprise and sad), are the same with the only active AU being 4. Tasks 1, 4, 5 (happy, startle, skeptical) also have the same highest frame count pattern with AUs 6, 7, 10, 12, and  14 being active. In DISFA, as can be seen in Table \ref{table:seqPatternDISFA}, the patterns with the highest frame count for each sequence are largely inactive AUs. Sequences 2, 3, and 5-9 only have 1 active AU, sequence 1 has 3 active AUs, and sequence 4 has 2 active AUs. The active AUs make up approximately 11\% of the AUs across the top patterns for each sequence.

Using action units, Lucey et al. \cite{lucey2010automatically} showed that they can recognize pain from the UNBC-McMaster shoulder pain expressive archive database \cite{lucey2011painful}. Although, these results are encouraging, analysis of AU patterns across multiple emotions suggests that using AUs to detect expression across a range (e.g. happy, sad) is a difficult problem due to similar AU patterns between each. This could explain, in part, the difficulty in detecting AUs as machine learning classifiers are getting trained on similar patterns, however, different expressions are used as training data. Considering this, we hypothesized that AU patterns have a direct impact on AU detection. We evaluate this hypothesis in the next section by training on patterns vs. training on single AUs.
\begin{table}
\centering
\captionsetup{justification=centering}
\caption{Top AU occurrence patterns, for each sequence, from DISFA. \textit{NOTE: Last column is target emotion of task.}}
\vspace{-2mm}
\newcolumntype{P}[1]{>{\centering\arraybackslash}p{#1}}
{\fontsize{6.5}{7}\selectfont
\begin{tabular}{|P{0.75cm}||P{0.09cm}|P{0.09cm}|P{0.09cm}|P{0.09cm}|P{0.09cm}|P{0.09cm}|P{0.09cm}|P{0.09cm}|P{0.09cm}|P{0.09cm}|P{0.09cm}|P{0.09cm}||P{0.7cm}| }
 \hline 
 \multirow{2}{*}{Seq} &\multicolumn{12}{c|}{Pattern} &\multirow{2}{*}{Emo}\\ \cline{2-13} &1 &2 &4 &5 &6 &9 &12 &15 &17 &20 &25 &26  &\\ \hline
 \rowcolor{lightgray}
1	&0 &0 &0 &0 &1 &0 &1 &0 &0 &0 &1 &0 &\\ \rowcolor{lightgray} \cline{1-13} \rowcolor{lightgray}
2	&0 &0 &0 &0 &0 &0 &1 &0 &0 &0 &0 &0 &\multirow{-2}{*}{Happy}\\ \hline 
3	&0 &0 &0 &0 &0 &0 &1 &0 &0 &0 &0 &0 &Surp\\ \hline 
\rowcolor{lightgray}
4	&0 &0 &0 &0 &0 &0 &0 &0 &0 &0 &1 &1 &Fear\\ \hline 
5	&0 &0 &0 &0 &0 &0 &0 &0 &0 &0 &1 &0 &\\ \cline{1-13} 
6	&0 &0 &0 &0 &0 &0 &0 &0 &0 &0 &1 &0 &\multirow{-2}{*}{Disg}\\ \hline 
\rowcolor{lightgray}
7	&0 &0 &0 &0 &0 &0 &0 &0 &0 &0 &1 &0 &\\ \cline{1-13}
\rowcolor{lightgray}
8	&0 &0 &1 &0 &0 &0 &0 &0 &0 &0 &0 &0 &\multirow{-2}{*}{Sad}\\ \hline 
9	&0 &0 &0 &0 &0 &0 &0 &0 &0 &0 &1 &0 &Surp\\ \hline 
 \end{tabular}
 }
\label{table:seqPatternDISFA}
\end{table}
\begin{table}
\centering
\captionsetup{justification=centering}
\caption{Top AU occurrence patterns, for each task, in BP4D. \textit{NOTE: Last column is target emotion of task.}}
\vspace{-2mm}
\newcolumntype{P}[1]{>{\centering\arraybackslash}p{#1}}
{\fontsize{6.5}{7}\selectfont
\begin{tabular}{|P{0.75cm}||P{0.09cm}|P{0.09cm}|P{0.09cm}|P{0.09cm}|P{0.09cm}|P{0.09cm}|P{0.09cm}|P{0.09cm}|P{0.09cm}|P{0.09cm}|P{0.09cm}|P{0.09cm}||P{0.7cm}| }
 \hline 
 \multirow{2}{*}{Task ID} &\multicolumn{12}{c|}{Pattern} &\multirow{2}{*}{Emo}\\ \cline{2-13} &1 &2 &4 &6 &7 &10 &12 &14 &15 &17 &23 &24  &\\ \hline
 
T1	&0 &0 &0 &1 &1 &1 &1 &1 &0 &0 &0 &0 &Happy\\ \hline 
\rowcolor{lightgray}
T2	&0 &0 &1 &0 &0 &0 &0 &0 &0 &0 &0 &0 &Surp\\ \hline 
T3	&0 &0 &1 &0 &0 &0 &0 &0 &0 &0 &0 &0 &Sad\\ \hline 
\rowcolor{lightgray}
T4	&0 &0 &0 &1 &1 &1 &1 &1 &0 &0 &0 &0 &Startle\\ \hline 
T5	&0 &0 &0 &1 &1 &1 &1 &1 &0 &0 &0 &0 &Skept\\ \hline 
\rowcolor{lightgray}
T6	&0 &0 &0 &1 &1 &1 &1 &0 &0 &0 &0 &0 &Emb\\ \hline 
T7	&0 &0 &0 &0 &0 &1 &1 &1 &0 &0 &0 &0 &Fear\\ \hline 
\rowcolor{lightgray}
T8	&0 &0 &0 &1 &1 &1 &1 &0 &0 &0 &0 &0 &Pain\\ \hline 
 \end{tabular}
 }
\label{table:seqPatternBP4D}
\vspace{-2mm}
\end{table}

\pgfplotstableread[col sep=&, header=true]{

Count           &BP4D           &DISFA
5000-11000	&0.11820331	    &1.515151515
2000-5000	    &0.059101655	&0.757575758
1000-2000	    &1.182033097	&2.272727273
500-1000	    &1.477541371	&3.787878788
200-500	    &5.082742317	&7.196969697
100-200	    &7.624113475	&7.954545455
50-100	        &11.5248227	    &10.60606061
10-50	        &33.27423168	&30.3030303
5-10	        &13.4751773	    &13.25757576
0-5 	        &26.1820331	    &22.34848485
}\dataPattern

\pgfplotstableread[col sep=&, header=true]{
AU	    &AUC	        &F1 Binary	    &F1 Micro	    &F1 Macro
AU 1	&0.669216343	&0.406828069	&0.679803744	&0.59018316
AU 2	&0.713818404	&0.379495718	&0.743093578	&0.606952745
AU 4	&0.708617001	&0.364931266	&0.771311218	&0.612599617
AU 6	&0.834497011	&0.745054337	&0.745827474	&0.745817826
AU 7	&0.746618837	&0.740872128	&0.679163149	&0.659074221
AU 10	&0.835096537	&0.817412506	&0.755584635	&0.7216427
AU 12	&0.887200044	&0.827086344	&0.795496095	&0.788324077
AU 14	&0.652125152	&0.622442648	&0.602801451	&0.600488803
AU 15	&0.66698335     &0.308669201	&0.739958515	&0.573713531
AU 17	&0.750157798	&0.597749615	&0.686180323	&0.669936602
AU 23	&0.690567254	&0.370420011	&0.780498789	&0.618559403
AU 24	&0.803408505	&0.336350097	&0.840115608	&0.622709823

}\matricgraph

\section{Impact of AU Patterns on Detection}
\label{sec:exp}
\pgfplotstableread[col sep=&, header=true]{
AU	    &Shallow	    &Deep	        &Ones	    
AU 1	&0.40682807	    &0.335677137	&0.35
AU 2	&0.37949572	    &0.347588473	&0.29
AU 4	&0.36493127	    &0.140961036	&0.34
AU 6	&0.74505434	    &0.471137237	&0.63
AU 7	&0.74087213	    &0.714831567	&0.71
AU 10	&0.81741251	    &0.781767707	&0.75
AU 12	&0.82708634	    &0.763914561	&0.72
AU 14	&0.62244265	    &0.629838631	&0.64
AU 15	&0.3086692	    &0.164031076	&0.29
AU 17	&0.59774962	    &0.413123734	&0.51
AU 23	&0.37042001	    &0.258251238	&0.28
AU 24	&0.3363501	    &0.240482167	&0.26

}\FBinary
\pgfplotstableread[col sep=&, header=true]{
AU	    &Shallow	    &Deep	        &Ones	    
AU 1	&0.67980374 	&0.629590717	&0.2
AU 2	&0.74309358	    &0.70118653	    &0.17
AU 4	&0.77131122	    &0.687293765	&0.2
AU 6	&0.74582747	    &0.654980119	&0.46
AU 7	&0.67916315	    &0.622762292	&0.55
AU 10	&0.75558463	    &0.697970564	&0.59
AU 12	&0.79549609	    &0.677751757	&0.56
AU 14	&0.60280145	    &0.563080642	&0.47
AU 15	&0.73995852	    &0.721819031	&0.17
AU 17	&0.68618032	    &0.575876067	&0.34
AU 23	&0.78049879	    &0.618210161	&0.17
AU 24	&0.84011561     &0.752592833	&0.15

}\FMicro

\pgfplotstableread[col sep=&, header=true]{
AU	    &Shallow	    &Deep	        &Ones	    
AU 1	&0.59018316	    &0.527118215	&0.174
AU 2	&0.606952745	&0.564804895	&0.15
AU 4	&0.612599617	&0.473916553	&0.17
AU 6	&0.745817826	&0.591707328	&0.31
AU 7	&0.659074221	&0.546353299	&0.35
AU 10	&0.7216427	    &0.610417486	&0.37
AU 12	&0.788324077	&0.610425783	&0.36
AU 14	&0.600488803	&0.508125217	&0.32
AU 15	&0.573713531	&0.497657889	&0.144
AU 17	&0.669936602	&0.524865512	&0.255
AU 23	&0.618559403	&0.497665599	&0.142
AU 24	&0.622709823	&0.546285511	&0.13

}\FMacro

\pgfplotstableread[col sep=&, header=true]{
AU	    &Shallow	    &Deep	        &Ones	    
AU 1	&0.669216343	&0.588047457	&0.5
AU 2	&0.713818404	&0.618745604	&0.5
AU 4	&0.708617001	&0.57614887	    &0.5
AU 6	&0.834497011	&0.699210252	&0.5
AU 7	&0.746618837	&0.628654392	&0.5
AU 10	&0.835096537	&0.712456826	&0.5
AU 12	&0.887200044	&0.737294786	&0.5
AU 14	&0.652125152	&0.583880398	&0.5
AU 15	&0.66698335	    &0.570476033	&0.5
AU 17	&0.750157798	&0.634941339	&0.5
AU 23	&0.690567254	&0.577412025	&0.5
AU 24	&0.803408505	&0.675005004	&0.5

}\AUC
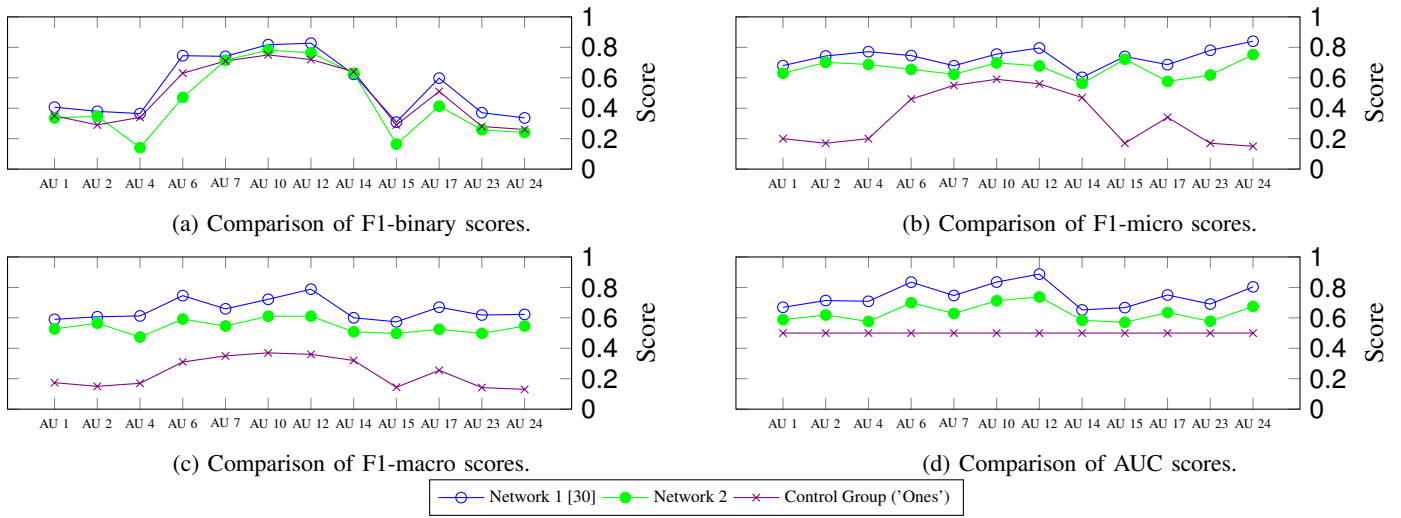
\begin{figure*}
\newenvironment{customlegend}[1][]{%
    \begingroup
    \csname pgfplots@init@cleared@structures\endcsname
    \pgfplotsset{#1}%
}{%
    \csname pgfplots@createlegend\endcsname
    \endgroup
}%

\def\addlegendimage{\csname pgfplots@addlegendimage\endcsname}
  \begin{subfigure}{0.5\linewidth}
\begin{tikzpicture}
 \centering
  
      \begin{axis}
      [
  ymin=0, ymax=1,
        xtick=data,
        xticklabels from table={\FBinary}{AU},      
        height=3.6cm, width=9cm,
        nodes near coords align={vertical},
        ylabel=Score,
         ylabel near ticks, 
         yticklabel pos=right,
         legend style={at={(0.5,-0.1)},anchor=north},
         legend columns=4,
         xticklabel style={font=\tiny} ,
      ]
 \addplot [mark=o,blue] table [y=Shallow, x expr=\coordindex] {\FBinary};
\addplot [mark=*,green] table [y=Deep, x expr=\coordindex] {\FBinary};
\addplot [mark=x,violet] table [y=Ones, x expr=\coordindex] {\FBinary};

\end{axis}

\end{tikzpicture}

    \caption{Comparison of F1-binary scores.}
  \label{graph:CompF1BinScore}
\end{subfigure}
\hspace{1em}
  \begin{subfigure}{0.5\linewidth}
\begin{tikzpicture}
 \centering
  
      \begin{axis}
      [
  ymin=0, ymax=1,
        xtick=data,
        xticklabels from table={\FMicro}{AU},      
        height=3.6cm, width=9cm,
        nodes near coords align={vertical},
         ylabel=Score,
         ylabel near ticks, 
         yticklabel pos=right,
         legend style={at={(0.5,-0.1)},anchor=north},
         legend columns=4,
         xticklabel style={font=\tiny} ,
      ]
  \addplot [mark=o,blue] table [y=Shallow, x expr=\coordindex] {\FMicro};
\addplot [mark=*,green] table [y=Deep, x expr=\coordindex] {\FMicro};
\addplot [mark=x,violet] table [y=Ones, x expr=\coordindex] {\FMicro};

\end{axis}

\end{tikzpicture}

    \caption{Comparison of F1-micro scores.}
  \label{graph:CompF1MicroScore}
\end{subfigure}
  \begin{subfigure}{0.5\linewidth}
\begin{tikzpicture}
 \centering
  
      \begin{axis}
      [
  ymin=0, ymax=1,
        xtick=data,
        xticklabels from table={\FMacro}{AU},      
        height=3.6cm, width=9cm,
        nodes near coords align={vertical},
         ylabel=Score,
         ylabel near ticks, 
         yticklabel pos=right,
         legend style={at={(0.5,-0.1)},anchor=north},
         legend columns=4,
         xticklabel style={font=\tiny} ,
      ]
  \addplot [mark=o,blue] table [y=Shallow, x expr=\coordindex] {\FMacro};
\addplot [mark=*,green] table [y=Deep, x expr=\coordindex] {\FMacro};
\addplot [mark=x,violet] table [y=Ones, x expr=\coordindex] {\FMacro};

\end{axis}
\end{tikzpicture}
    \caption{Comparison of F1-macro scores.}
  \label{graph:CompF1MacroScore}
\end{subfigure}
 \hspace{1em}
  \begin{subfigure}{0.5\linewidth}
\begin{tikzpicture}
 \centering
  
      \begin{axis}
      [
  ymin=0, ymax=1,
        xtick=data,
        xticklabels from table={\AUC}{AU},      
        height=3.6cm, width=9cm,
        nodes near coords align={vertical},
         ylabel=Score,
         ylabel near ticks, 
         yticklabel pos=right,
         legend style={at={(0.5,-0.1)},anchor=north},
         legend columns=4,
         xticklabel style={font=\tiny} ,
      ]
  \addplot [mark=o,blue] table [y=Shallow, x expr=\coordindex] {\AUC};
\addplot [mark=*,green] table [y=Deep, x expr=\coordindex] {\AUC};
\addplot [mark=x,violet] table [y=Ones, x expr=\coordindex] {\AUC};

\end{axis}

\end{tikzpicture}

    \caption{Comparison of AUC scores.}
  \label{graph:CompF1AUScore}
\end{subfigure}

\centering
\begin{tikzpicture}
{\fontsize{6.5}{7}\selectfont
    \begin{customlegend}[legend columns=-1, legend entries={Network 1 \cite{ertugrulFG2019}, Network 2, Control Group ('Ones')}]
    \addlegendimage{blue,mark=o}
    \addlegendimage{green,mark=*}
    \addlegendimage{violet,mark=x}
    \end{customlegend}
    }
\end{tikzpicture}
  \caption{Comparisons of different accuracy metrics on multiple networks (\textit{experiment 1}).}
  \label{graph:OriginalCompare}
  \vspace{-4mm}
\end{figure*}
To further investigate the impact of AU occurrence patterns on detection, we conducted in-depth experiments on BP4D. We chose BP4D for our experiments as DISFA contains a larger imbalance of active versus inactive AUs \cite{li2018eac}. We evaluated the impact of two different convolutional neural networks (CNN) for this investigation to validate that the results are not classifier (i.e. network) specific. First, we implemented the CNN as detailed by Ertugral et al.\cite{ertugrulFG2019}, which we refer to as \textit{Network 1}. For our second CNN, we used a network which had two convolutional layers with filter size of 8 and 16 followed by max pool layers, followed by two more convolutional layers, with filter sizes of 16 and 20; another max pool layer and batch normalization. All CNNs used had a kernel of (3,3). There were three dense layers, before the output layer, with 4096, 4096 and 512 neurons respectively, relu activation function was used and dropout of 0.4. We refer to this as \textit{Network 2}. In addition to these two networks, we also had a control group called 'Ones', in which we labeled all AUs as active in all the frames, and is used as a baseline for comparisons between the two networks.

\begin{table}
\centering
\captionsetup{justification=centering}
\caption{Correlation between metrics and class imbalance for \textit{experiment 1}.}
\vspace{-2mm}
\newcolumntype{P}[1]{>{\centering\arraybackslash}p{#1}}
{\fontsize{6.5}{8}\selectfont
\begin{tabular}{ |P{1.5cm}||P{1.5cm}|P{1.5cm}|P{1cm}|  }
 \hline 
 \multirow{2}{*}{Metric} 	&\multicolumn{3}{c|}{Correlation}\\ \cline{2-4} &Network 1 &Network 2 &Ones \\  \hline 
	
F1-binary	&0.9758     &0.9489     &0.9912 \\ \hline 
F1-macro	&0.7570     &0.6349     &0.9961 \\ \hline 
AUC         &0.5795     &0.6406     &N/A  \\ \hline 
F1-micro	&-0.2656   &-0.3118     &0.9913 \\ \hline 
 \end{tabular}
 }
\label{table:Org Metric of Corrl}
\vspace{-7mm}
\end{table}
To investigate the impact of the AU patterns on detection accuracy, we calculated the F1-binary, F1-micro, F1-macro, and Area Under the Curve (AUC) scores. F1 score is defined as $F1 = \frac{2\times True Positives}{2\times True Positives + False Positives + False Negatives}$ \cite{chase2014thresholding} and AUC is defined as the area under the graph between True positive rate V/S False positive rate. F1-binary is F1 score for the positive class and does not consider the negative class where as F1-macro is the simple average of F1 scores of all classes. F1-micro is calculated globally by counting the total true positives, and false positives and negatives. The score is computed using all estimated labels and manual annotations without averaging over the folds. To facilitate our investigation, we conducted two experiments. First, using the entire dataset, we detected multiple AUs (i.e. the entire sequence/pattern). For this experiment, we used all AU-labeled frames from BP4D, and we refer to this as \textit{experiment 1}. Secondly, we detected individual AUs, which we refer to this as \textit{experiment 2} in the rest of the paper, and was done to test what impact removing the patterns has on AU the detection accuracy. Both experiments were subject-independent (i.e. same subject does not appear in training and testing), and the subjects in the each fold were fixed so that both experiments trained and tested the same subjects and images. Three-fold cross-validation was used for all experiments.

\subsection{Multi-AU detection (Experiment 1)}
When using the 'Ones' control group as a baseline, it can be seen that there is a high correlation between the class imbalance and F1-binary, macro, and micro scores (Table \ref{table:Org Metric of Corrl}). There is an average correlation, with the class imbalance, of 0.9928 across the three metrics. While the accuracies vary between the different metrics, it can be seen that the trend is similar (Fig. \ref{graph:OriginalCompare}). For the control group, the AUC correlation is N/A as a score of 0.5 was obtained for each AU (Fig. \ref{graph:CompF1AUScore}). 

As can be seen in Fig. \ref{graph:CompF1BinScore}, when the F1-binary scores of the tested networks is similar to the control group. It can be seen that all of them follow a similar trend, which is the class imbalance. This suggests that the F1-binary score may not be an accurate metric to distinguish between correct detection and guessing (i.e. "guessing" all AUs as ones/active). This can be explained, in part, since the F1-binary score only looks at the positive classes \cite{chase2014thresholding}. This can also be seen in Table \ref{table:Org Metric of Corrl} (first row), as there is a high correlation between the F1-binary score of both networks and the class imbalance. We also calculated the correlation across each AU of both networks to the control group (i.e. how correlated are the F1-binary accuracies for each AU). This resulted in correlations of 0.98 and 0.94 for \textit{Networks 1 and 2}, respectively, showing both give similar results to labeling all AUs as active.

We also looked at using F1-micro and macro as the metrics for AU detection accuracy. F1-micro does not follow the control group trend. It has a correlation of -0.29 and -0.31 for \textit{Networks 1 and 2}, respectively, with the control group (Fig. \ref{graph:CompF1MicroScore}). It also had a low negative correlation with the class imbalance for both networks (Table \ref{table:Org Metric of Corrl}). F1-macro also does not follow the control group trend (Fig. \ref{graph:CompF1MacroScore}). Although F1-macro is more correlated compared to F1-micro, with correlations of 0.74 and 0.62 for \textit{Networks 1 and 2}, respectively, it is less correlated compared to F1-binary (Table \ref{table:Org Metric of Corrl}).

\pgfplotstableread[col sep=&, header=true]{
AU	    &AUC	        &Binary	        &Micro	        &Macro
AU 1	&0.685701086	&0.590648787	&0.628401971	&0.621103458
AU 2	&0.602277835	&0.497334225	&0.584498134	&0.568534803
AU 4	&0.695899879	&0.504281976	&0.633439674	&0.606739374
AU 6	&0.833332452	&0.749139464	&0.744542496	&0.744403376
AU 7	&0.765387571	&0.730914615	&0.692071751	&0.683720695
AU 10	&0.809489313	&0.742660392	&0.714612828	&0.70894957
AU 12	&0.901637949	&0.825001733	&0.821194523	&0.819679761
AU 14	&0.632881881	&0.596056026	&0.587730868	&0.586324067
AU 15	&0.715063926	&0.565229487	&0.644930801	&0.630271392
AU 17	&0.724073693	&0.653666434	&0.647716801	&0.645276524
AU 23	&0.661369833	&0.507801303	&0.622707073	&0.596589613
AU 24	&0.801232345	&0.580642488	&0.679997875	&0.658587816

}\BalShallow

\pgfplotstableread[col sep=&, header=true]{
AU	    &Binary	    &Micro	&Macro	&AUC
AU 1	&0.438	    &0.494	&0.331	&0.5
AU 2	&0.439	    &0.502	&0.333	&0.5
AU 4	&0.223	    &0.6	&0.376	&0.5
AU 6	&0.21	    &0.488	&0.327	&0.5
AU 7	&0.432	    &0.457	&0.324	&0.5
AU 10	&0.218	    &0.502	&0.334	&0.5
AU 12	&0.443	    &0.489	&0.329	&0.5
AU 14	&0.452	    &0.51	&0.335	&0.5
AU 15	&0.212	    &0.513	&0.339	&0.5
AU 17	&0.414	    &0.482	&0.332	&0.5
AU 23	&0.201	    &0.5233	&0.346	&0.5
AU 24	&0.423	    &0.502	&0.33	&0.5

}\Baldeep
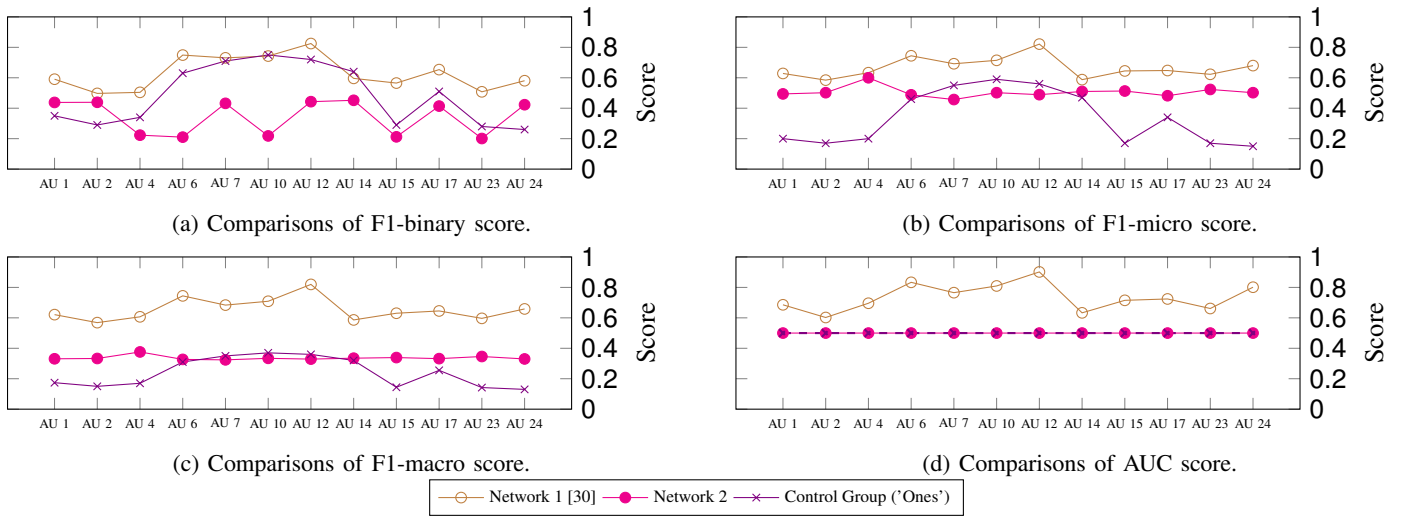
\begin{figure*}
\newenvironment{customlegend}[1][]{%
    \begingroup
    \csname pgfplots@init@cleared@structures\endcsname
    \pgfplotsset{#1}%
}{%
    \csname pgfplots@createlegend\endcsname
    \endgroup
}%

\def\addlegendimage{\csname pgfplots@addlegendimage\endcsname}
  \begin{subfigure}{0.5\linewidth}
\begin{tikzpicture}
 \centering
  
      \begin{axis}
      [
  ymin=0, ymax=1,
        xtick=data,
        xticklabels from table={\BalShallow}{AU},      
        height=3.6cm, width=9cm,
        nodes near coords align={vertical},
         ylabel=Score,
         ylabel near ticks, 
         yticklabel pos=right,
         legend style={at={(0.5,-0.1)},anchor=north},
         legend columns=4,
         xticklabel style={font=\tiny} ,
      ]
  \addplot [mark=o,brown] table [y=Binary, x expr=\coordindex] {\BalShallow};
\addplot [mark=*,magenta] table [y=Binary, x expr=\coordindex] {\Baldeep};
\addplot [mark=x,violet] table [y=Ones, x expr=\coordindex] {\FBinary};

\end{axis}

\end{tikzpicture}

    \caption{Comparisons of F1-binary score.}
  \label{graph:BalF1BinScore}
\end{subfigure}
\hspace{1em}
  \begin{subfigure}{0.5\linewidth}
\begin{tikzpicture}
  
      \begin{axis}
      [
  ymin=0, ymax=1,
        xtick=data,
        xticklabels from table={\FMicro}{AU},      
        height=3.6cm, width=9cm,
        nodes near coords align={vertical},
         ylabel=Score,
         ylabel near ticks, 
         yticklabel pos=right,
         legend style={at={(0.5,-0.1)},anchor=north},
         legend columns=4,
         xticklabel style={font=\tiny} ,
      ]
  \addplot [mark=o,brown] table [y=Micro, x expr=\coordindex] {\BalShallow};
\addplot [mark=*,magenta] table [y=Micro, x expr=\coordindex] {\Baldeep};
\addplot [mark=x,violet] table [y=Ones, x expr=\coordindex] {\FMicro};

\end{axis}

\end{tikzpicture}

    \caption{Comparisons of F1-micro score.}
  \label{graph:BalF1MicroScore}
\end{subfigure}
  \begin{subfigure}{0.5\linewidth}
\begin{tikzpicture}
 \centering
  
      \begin{axis}
      [
  ymin=0, ymax=1,
        xtick=data,
        xticklabels from table={\FMacro}{AU},      
        height=3.6cm, width=9cm,
        nodes near coords align={vertical},
         ylabel=Score,
         ylabel near ticks, 
         yticklabel pos=right,
         legend style={at={(0.5,-0.1)},anchor=north},
         legend columns=4,
         xticklabel style={font=\tiny} ,
      ]
  \addplot [mark=o,brown] table [y=Macro, x expr=\coordindex] {\BalShallow};
\addplot [mark=*,magenta] table [y=Macro, x expr=\coordindex] {\Baldeep};
\addplot [mark=x,violet] table [y=Ones, x expr=\coordindex] {\FMacro};

\end{axis}
\end{tikzpicture}
    \caption{Comparisons of F1-macro score.}
  \label{graph:BalF1MacroScore}
\end{subfigure}
 \hspace{1em}
  \begin{subfigure}{0.5\linewidth}
\begin{tikzpicture}
 \centering
  
      \begin{axis}
      [
  ymin=0, ymax=1,
        xtick=data,
        xticklabels from table={\AUC}{AU},      
        height=3.6cm, width=9cm,
        nodes near coords align={vertical},
         ylabel=Score,
         ylabel near ticks, 
         yticklabel pos=right,
         legend style={at={(0.5,-0.1)},anchor=north},
         legend columns=4,
         xticklabel style={font=\tiny} ,
      ]
  \addplot [mark=o,brown] table [y=AUC, x expr=\coordindex] {\BalShallow};
\addplot [mark=*,magenta] table [y=AUC, x expr=\coordindex] {\Baldeep};
\addplot [mark=x,thick,dashed,violet] table [y=Ones, x expr=\coordindex] {\AUC};

\end{axis}

\end{tikzpicture}

    \caption{Comparisons of AUC score.}
  \label{graph:BalF1AUCScore}
\end{subfigure}

\centering
\begin{tikzpicture}
{\fontsize{6.5}{7}\selectfont
    \begin{customlegend}[legend columns=-1,  legend entries={Network 1 \cite{ertugrulFG2019}, Network 2, Control Group ('Ones')}]
    \addlegendimage{brown,mark=o}
    \addlegendimage{magenta,mark=*}
    \addlegendimage{violet,mark=x}
    \end{customlegend}
    }
\end{tikzpicture}
  \caption{Comparisons of different accuracy metrics on different networks (\textit{experiment 2}). \textit{NOTE: Best viewed in color.}}
  \label{graph:balancedDataComparisons}
  \vspace{-4mm}
  \end{figure*}
  \begin{table}
\centering
\captionsetup{justification=centering}
\caption{Correlation between metrics and class imbalance for \textit{experiment 2}.}
\vspace{-2mm}
\newcolumntype{P}[1]{>{\centering\arraybackslash}p{#1}}
{\fontsize{6.5}{7.5}\selectfont
\begin{tabular}{ |P{1.5cm}||P{1.5cm}|P{1.5cm}|  }
 \hline 
 \multirow{2}{*}{Metric} 	&\multicolumn{2}{c|}{Correlation}\\ \cline{2-3} &Network 1 &Network 2\\  \hline 
	
F1-binary	&0.8783    &0.1121     \\ \hline 
F1-macro	&0.6869     & -0.4281   \\ \hline 
AUC         &0.5840     &N/A     \\ \hline 
F1-micro	&0.6290   &-0.4679    \\ \hline 
 
 \end{tabular}
 }

\label{table:Bal Metric of Corrl}
\vspace{-5mm}
\end{table}
The final metric we looked at, for AU detection accuracy, was AUC. This metric has a lower correlation, with the class imbalance, compared to F1-macro. It can also be seen that it does not follow the class imbalance trend (Fig. \ref{graph:CompF1AUScore}). Again, as AUC for this experiment was 0.5, we were unable to calculate the correlation between the control group and two networks. It is also important to note that the correlations between the control group and F1-binary, macro, and micro closely resemble the correlation with the class imbalance. This is due to the high correlation of the control group with the data. 

\subsection{Single-AU Detection (Experiment 2)}
In \textit{experiment 1}, we showed that class imbalance has a direct impact on AU detection, as some of the evaluation metrics are similar to manually labeling all AUs as 1. To validate our hypothesis that action unit occurrence patterns have a direct impact on AU detection we trained separate networks for each individual AU (i.e. single AU detection). This experimental design allows us to directly compare the correlations, and evaluation metrics obtained from \textit{experiment 1}. Similar to \textit{experiment 1}, we calculated the correlations between the F1-binary, macro, micro and AUC scores compared to the class imbalance for the two networks (Table \ref{table:Bal Metric of Corrl}). 

In \textit{experiment 1}, the correlations with the class imbalance, for each metric, across the two network architectures was similar (e.g. F1-binary correlation of 0.9758 and 0.9489 for \textit{Networks 1 and 2}, respectively). Conversely, for \textit{experiment 2}, the correlations, between the two networks, are not similar. For example, the correlation between F1-micro for \textit{Network 1} is 0.6290, while \textit{Network 2} has a correlation of -0.4679. Similar to \textit{experiment 1}, the AUC for our control group was 0.5 for all AUs, however, the AUC for \textit{Network 2} was also 0.5 for all AUs (Fig. \ref{graph:BalF1AUCScore}). This can be explained, in part, by the general performance of \textit{ Network 2} as seen in Fig. \ref{graph:balancedDataComparisons}. Each metric, for Network 2, generally performed poorly, compared to \textit{experiment 1} (i.e. multi-AU detection), which supports our hypothesis that AU occurrence patterns have a direct impact on AU detection. Our results suggest that the classifiers (i.e. \textit{Networks 1 and 2}), are learning the patterns of AU occurrences and not 1 AU at a time. This is validated through \textit{experiment 2}, that shows accuracies decrease when patterns are removed from training. Although the overall trend is that accuracies decrease when patterns are removed, it is interesting to note that AUs with lower base rates can be negatively impacted as seen in Figs. \ref{graph:OriginalCompare} and \ref{graph:balancedDataComparisons}. It has been shown that AU co-occurrences are important \cite{song2015exploiting}, \cite{zhao2015joint}, however, co-occurrence matrices are generally shown between 2 AUs \cite{mavadati2016extended}. Here, we take that one step further and show that the entire patterns of active and inactive AU occurrences directly impacts AU detection. Our results suggest that higher AU scores can be achieved with a multi-AU detection approach, which validates other work that has shown the same thing \cite{li2017action}.

\section{Training on AU Occurrence Patterns}
\label{sec:trainPat}
We propose a method that uses AU occurrence patterns to train deep neural networks to increase AU detection accuracies. To facilitate this, we conducted experiments on BP4D, as DISFA contains a larger imbalance of active versus inactive AUs \cite{li2018eac} and it has more patterns with fewer active AUs (Table \ref{table:patternDISFA}). We conducted two experiments; in the first experiment we created a subset of the dataset using the top 66 patterns, which we refer to as \textit{experiment 3}. In the second experiment, we used all patterns, which we refer to as \textit{experiment 4}. 

\begin{table*}[h]
  \centering
  \caption{Evaluation on tested networks for top 66 AU occurrence patterns.}
  \vspace{-2mm}
  \resizebox{\textwidth}{!}{%
    \begin{tabular}{|c|c|c|c|c|c|c|c|c|c|c|c|c|c|c|c|c|c|c|c|c|}
    \hline
          & \multicolumn{5}{c|}{F1 Bin} & \multicolumn{5}{c|}{AUC} & \multicolumn{5}{c|}{F1 Micro} & \multicolumn{5}{c|}{F1 Macro} \\
    \hline
    AU    & N1    & N3    & N4     &N5     &N6    & N1    & N3    & N4     &N5     &N6      & N1    & N3    & N4     &N5     &N6   & N1    & N3    & N4     &N5     &N6\\
    \hline
    1  & 0.2085 & 0.1982 & 0.2154 & 0.1556 & 0.2872 & 0.6444 & 0.5630 & 0.599 & 0.6467 & 0.6469 & 0.6797 & 0.8025 & 0.8323 & 0.8705 & 0.8314 & 0.5900 & 0.5587 & 0.5760 & 0.5426 & 0.5957\\
    \hline
    2  & 0.2571 & 0.2428  & 0.2666  & 0.1569 & 0.2633 & 0.6917 & 0.5757 & 0.7040 & 0.5972 & 0.6425 & 0.7432 & 0.8373 & 0.8690 & 0.886 & 0.8396 & 0.6070 & 0.5690 & 0.6040 & 0.5477 & 0.5864 \\
    \hline
    4  & 0.3238 & 0.2057 & 0.2331 & 0.0952 & 0.3433 & 0.7857 & 0.5600 & 0.6957 & 0.6257 & 0.7115 & 0.7712 & 0.8337 & 0.8613 & 0.8449 & 0.8233 & 0.6126 & 0.5630 & 0.6090 & 0.5052 & 0.6206 \\
    \hline
    6  & 0.8440 & 0.7931 & 0.8267 & 0.8104 & 0.7919 & 0.9095 & 0.8000 & 0.9027 & 0.7899 & 0.858 & 0.7460 & 0.8032 & 0.8380 & 0.7933 & 0.7883 & 0.7459 & 0.7997 & 0.7536 & 0.7904 & 0.7879 \\
    \hline
    7  & 0.7581 & 0.7409 & 0.7493 & 0.7406 & 0.7495  & 0.8037 & 0.7249 & 0.8037 & 0.7175 & 0.7488 & 0.6792 & 0.7289 & 0.7550 & 0.7055 & 0.7211 & 0.6593 & 0.7236 & 0.6613 & 0.6993 & 0.7174\\
    \hline
    10 & 0.9010 & 0.8641 & 0.8753 & 0.8732 & 0.8684 & 0.9442 & 0.8283 & 0.9504 & 0.7868 & 0.9008 & 0.7555 & 0.8460 & 0.8673 & 0.8406 & 0.8408 & 0.7219 & 0.8340 & 0.7477 & 0.8278 & 0.8315\\
    \hline
    12 & 0.9095 & 0.8758 & 0.8724  & 0.8886 & 0.8842 & 0.9631 & 0.8543 & 0.9484 & 0.8467 & 0.9307 & 0.7956 & 0.8673 & 0.8867 & 0.8627 & 0.8614 & 0.7883 & 0.8593 & 0.8137 & 0.8531 & 0.855 \\
    \hline
    14 & 0.5665 & 0.5134 & 0.5919 & 0.545 & 0.5269 & 0.6724 & 0.6167 & 0.6780 & 0.6749 & 0.6381 & 0.6027 & 0.6202 & 0.6277 & 0.6033 & 0.6103 & 0.6003 & 0.6120 & 0.5843 & 0.5967 & 0.5974\\
    \hline
    15 & 0.1668 & 0.1378 & 0.1633 & 0.0939 & 0.1643 & 0.7463 & 0.5680 & 0.7469 & 0.5713 & 0.6934 & 0.7397 & 0.8497 & 0.9103 & 0.9163 & 0.8672 & 0.5737 & 0.5430 & 0.5757 & 0.525 & 0.5459 \\
    \hline
    17 & 0.4357 & 0.3574 & 0.4241 & 0.2422 & 0.3224 & 0.7619 & 0.6097 & 0.7699 & 0.591 & 0.6205 & 0.6860 & 0.7703 & 0.8253 & 0.7995 & 0.7564 & 0.6700 & 0.6060 & 0.6483 & 0.5632 & 0.5866\\
    \hline
    23 & 0.2370 & 0.1118 & 0.2286 & 0.0422 & 0.1464  & 0.6499 & 0.5643 & 0.6680 & 0.5677 & 0.6181 & 0.7807 & 0.8982 & 0.9483 & 0.9445 & 0.9151 & 0.6184 & 0.5406 & 0.6050 & 0.5068 & 0.5508\\
    \hline
    24 & 0.0301 & 0.2147 & 0.2610 & 0.0132 & 0.0958 & 0.7712 & 0.5250 & 0.8117 & 0.528 & 0.6441 & 0.8402 & 0.9287 & 0.9479 & 0.9396 & 0.9285 &0.6227 & 0.5317 & 0.6143 & 0.491 & 0.5293\\
    \hline
    Avg & 0.4698 & 0.4380 & 0.4756 & 0.491 & 0.5293  & 0.7787 & 0.6492 & 0.7772 & 0.658 & 0.7211 & 0.7350 & 0.8155 & 0.8474 & 0.8339 & 0.8153 & 0.6508 & 0.6450 & 0.6494 & 0.6207 & 0.6504\\
    \hline
    \end{tabular}%
    }
  \label{tab:top66}%
  \vspace{-4mm}
\end{table*}%
Following the experimental design from Section \ref{sec:exp}, subject-independent, 3-fold cross validation was used where the subjects in each fold were fixed so that all the experiments were trained and tested on the same subjects. 

\subsection{Neural Network Architectures}
\label{sec:archs}
For training neural networks on AU occurrence patterns, we trained six networks. First, we trained the network from Ertugrul et al. \cite{ertugrulFG2019} (\textit{Network 1}). Secondly, we modified the network by changing the output layer from 12 to 66, to detect individual patterns. In this network we used class labels $[0,65]$, with one class for each pattern. For example, given the pattern where all AUs are 0, this pattern would be labeled with a class of 0. We also changed the loss function to categorical cross entropy, and the monitoring metric to accuracy. We refer to this network as \textit{Network 3}. Next, we froze the CNN layers from \textit{Network 3} and then split the network after these layers and added a 400 neuron fully connected layer followed by a 12 neuron output layer. To train this network we followed the same parameters as \textit{Network 1}. We refer to this network as \textit{Network 4}. To further validate the proposed approach, we also trained the network described by Fung et al. \cite{fung2019scalable}, which we refer to as \textit{Network 5}. We then made the same modifications as done to \textit{Network 1}, to create \textit{Networks 3 and 4}, on \textit{Network 5}. We changed the output layer from 12 to 66 to detect individual patterns for \textit{Network 6}, and we froze the CNN layers of \textit{Network 6} and then split the network after these layers for \textit{Network 7} to detect AUs. 

\subsection{Top 66 AU Occurrence Patterns (Experiment 3)}
\label{sec:top66}
For training we initially used the top 25 occurring patterns but found that AU 23 does not occur in the top 25 patterns and others like AU 2 and AU 15 occur very few number of times for reliable training and testing. For reliable training and testing of all AU occurrences, we used patterns which occur at least 397 times in the dataset, this gave us 66 top patterns. For this experiment, we trained \textit{Networks 1, 3, 4, 5, 6 and 7}.

As can be seen from Table \ref{tab:top66}, detecting AUs by first training on AU occurrence patterns (\textit{Network 4}) has the highest average F1-binary and F1-micro scores compared to \textit{Networks 1 and 3}. This shows that the proposed method has a positive impact on detection accuracy. An interesting result is that \textit{Network 1} has an F1-binary score of 0.03141 for AU 24, while the proposed approach has an F1-score of 0.261 for AU 24. This could be partially explained by the lack of training data (i.e. only top 66 patterns). \textit{Network 1} was unable to accurately detect this particular AU but the proposed approach was able to increase the accuracy with the limited amount of available training data. As \textit{Network 1} is from Ertugrul et al. \cite{ertugrulFG2019}, this validates that the proposed approach can be integrated into current state-of-the-art networks to further improve upon results. This can also be seen in Table \ref{tab:var}, where the proposed approach also has the lowest variance, across F1 scores of all AUs, for \textit{experiments 1 and 2}. As also can be seen in Table \ref{tab:top66}, \textit{Network 6} has higher average F1-binary, AUC, and F1-macro scores compared to \textit{Network 5}. Along with this, \textit{Network 7} has an average F1-binary, AUC, and F1-micro and -macro score of 0.4055, 0.7389, 0.8443, and 0.6225, respectively. Compared to \textit{Networks 5 and 6}, that is a higher AUC and F1-micro which shows similar results when comparing \textit{Network 4} to \textit{Networks 1 and 3}. These results show the utility of the proposed approach to help mitigate the negative impact that the imbalance of AU occurrence patterns have on detection evaluation metrics.

We are also interested in what would happen if we were given an unseen pattern to detect. To facilitate this, we trained \textit{Network 3} on the top 66 patterns, and then tested on all other available patterns. As can be seen in Table \ref{tab:unseen}, there is a decrease in overall scores, however, the proposed approach was still able to detect the unseen patterns with a relatively low decrease in accuracy. For example, when detecting unseen patterns the F1-binary score was 0.4361, however, when training and testing only on the top 66 patterns, with \textit{Network 3}, the F1-binary score was 0.4380 which is a decrease of 0.0019 in F1-binary score. It has been shown that cross-domain AU detection is a difficult problem \cite{ertugrul2020crossing}, which could explain, in part, the decrease in accuracy. Testing unseen patterns can be seen as an out of distribution problem which is similar to what is seen in cross-domain AU detection.

\subsection{All AU Occurrence Patterns (Experiment 4)}
\label{sec:allPatterns}
Here, instead of the top 66 patterns, we used all patterns from BP4D. We trained \textit{Networks 1, 4, 5, and 7} with the same CNN weights as we did in Section \ref{sec:archs}. For this experiment, we did not train \textit{Networks 3 and 6}, as there were not enough instances of all the patterns. As can be seen in Table \ref{tab:AllAU}, the average F1 score, AUC, and F1-micro of \textit{Network 4} are higher than \textit{Network 1}. For \textit{Network 7}, the average F1 score, AUC, F1-micro and macro are higher than \textit{Network 5}. These results further validate that the proposed approach can be integrated into current networks \cite{ertugrulFG2019, fung2019scalable} to help improve state-of-the-art results. Similar to \textit{experiment 3}, as also can be seen in Table \ref{tab:var}, the method of pre-training the CNN with the patterns has decreased the variance of the F1 scores.
\begin{table*}
  \centering
  \caption{Evaluation on tested networks for all AU occurrence patterns.}
  \vspace{-2mm}
{\fontsize{6}{5}\selectfont
    \begin{tabular}{|c|c|c|c|c|c|c|c|c|c|c|c|c|c|c|c|c|}
    \hline
          & \multicolumn{4}{c|}{F1 Bin} & \multicolumn{4}{c|}{AUC} & \multicolumn{4}{c|}{F1 Micro} & \multicolumn{4}{c|}{F1 Macro} \\
    \hline
    AU    & N1    & N4    &N5    &N7    & N1    & N4    &N5    &N7   & N1    & N4    &N5    &N7    & N1    & N4    &N5    &N7 \\
    \hline
    1     & 0.4068 & 0.4339 & 0.2137 & 0.3579 & 0.6250 & 0.6693 & 0.5876 & 0.6684 & 0.6797 & 0.6813 & 0.7632 & 0.7681 & 0.5900 & 0.5760 & 0.5359 & 0.6081 \\
    \hline
    2     & 0.3795 & 0.3889 & 0.2054 & 0.2816 & 0.6553 & 0.7140 & 0.5813 & 0.6217 & 0.7432 & 0.7613 & 0.8108 & 0.7752 & 0.6070 & 0.6040 & 0.5475 & 0.5737 \\
    \hline
    4     & 0.3649 & 0.3775 & 0.1749 & 0.226  & 0.6896 & 0.7086 & 0.5937 & 0.6377 & 0.7712 & 0.7710 & 0.7617 & 0.7571 & 0.6126 & 0.6090 & 0.5166 & 0.5409 \\
    \hline
    6     & 0.7450 & 0.7085 & 0.4625 & 0.6899 & 0.8293 & 0.8343 & 0.6777 & 0.7772 & 0.7460 & 0.7543 & 0.6365 & 0.6952 & 0.7459 & 0.7536 & 0.5714 & 0.6940 \\
    \hline
    7     & 0.7408 & 0.6840 & 0.7026 & 0.7125 & 0.7187 & 0.7467 & 0.6142 & 0.7087 & 0.6792 & 0.6717 & 0.5988 & 0.6621 & 0.6593 & 0.6613 & 0.5191 & 0.6474 \\
    \hline
    10    & 0.8174 & 0.7978 & 0.7869 & 0.7772 & 0.8140 & 0.8351 & 0.6937 & 0.7683 & 0.7555 & 0.7777 & 0.6849 & 0.7027 & 0.7219 & 0.7477 & 0.5706 & 0.6584 \\
    \hline
    12    & 0.8270 & 0.7929 & 0.7622 & 0.7974 & 0.8845 & 0.8872 & 0.7219 & 0.8395 & 0.7956 & 0.8197 & 0.6713 & 0.7556 & 0.7883 & 0.8137 & 0.5978 & 0.7422 \\
    \hline
    14    & 0.6224 & 0.6023 & 0.3963 & 0.5600 & 0.6300 & 0.6521 & 0.564  & 0.5850 & 0.6027 & 0.5860 & 0.5533 & 0.5648 & 0.6003 & 0.5843  & 0.5533 & 0.5648 \\
    \hline
    15    & 0.3087 & 0.3485 & 0.1214 & 0.2143 & 0.6460 & 0.6671 & 0.5581 & 0.6213 & 0.7397 & 0.7573 & 0.7979 & 0.7946 & 0.5737 & 0.5757 & 0.4867 & 0.5480 \\
    \hline
    17    & 0.5977 & 0.6203 & 0.3192 & 0.4789 & 0.7107 & 0.7501 & 0.6084 & 0.6494 & 0.6860 & 0.6730 & 0.6271 & 0.6386 & 0.6700 & 0.6483 & 0.5242 & 0.5989 \\
    \hline
    23    & 0.3704 & 0.4105 & 0.1161 & 0.2233 & 0.6440 & 0.6904 & 0.5681 & 0.6199 & 0.7807 & 0.7987 & 0.794  & 0.7930 & 0.6184 & 0.6050 & 0.4993 & 0.5515 \\
    \hline
    24    & 0.3363 & 0.3763 & 0.1412 & 0.1376 & 0.7450 & 0.8034 & 0.6536 & 0.7283 & 0.8402 & 0.8413 & 0.8271 & 0.8346 & 0.6227 & 0.6143 & 0.5389 & 0.5229 \\
    \hline
    Avg & 0.5431 & 0.5451 & 0.3669 & 0.4547 & 0.7160 & 0.7465 & 0.6185 & 0.6854 & 0.7350 & 0.7411 & 0.7105 & 0.7285 & 0.6508 & 0.6494 & 0.5329 & 0.6041 \\
    \hline
    \end{tabular}}%
  \label{tab:AllAU}%
  \vspace{-2mm}
\end{table*}%

\begin{table}
  \centering
  \caption{Evaluation on \textit{N3} for unseen patterns.}
  \vspace{-2mm}
  {\fontsize{6.5}{7}\selectfont
    \begin{tabular}{|c|c|c|c|c|c|c|c|c|}
    \hline
       AU   & {F1 Bin} & {AUC} & {F1 Micro} & {F1 Macro} \\
    \hline
    
    \hline
    1     & 0.2763 & 0.5480  & 0.6570 & 0.5250 \\
    \hline
    2     & 0.2443 & 0.5500  & 0.7171  & 0.5277 \\
    \hline
    4     & 0.1950 & 0.5270  & 0.6829 & 0.4988 \\
    \hline
    6     & 0.6433 & 0.6590  & 0.6468  & 0.6453 \\
    \hline
    7     & 0.6453  & 0.5876  & 0.5950  & 0.5857 \\
    \hline
    10    & 0.7367  & 0.6377 & 0.6673 & 0.6367 \\
    \hline
    12    & 0.7280 & 0.6850 & 0.6913 & 0.6827 \\
    \hline
    14    & 0.5663 & 0.5483 & 0.5457 & 0.5430 \\
    \hline
    15    & 0.2693  & 0.5337  & 0.6577  & 0.5224 \\
    \hline
    17    & 0.4855  & 0.5780 & 0.5620 & 0.5493 \\
    \hline
    23    & 0.2483  & 0.5320  & 0.6557 & 0.5123 \\
    \hline
    24    & 0.1947  & 0.5384  & 0.7277 & 0.5147 \\
    \hline
    Average  & 0.4361  & 0.5771  & 0.6505  & 0.5620 \\
    \hline
    \end{tabular}%
    }
  \label{tab:unseen}%
\end{table}%

\begin{table}
  \centering
  \caption{Variance of AU F1-binary scores.}
  \vspace{-2mm}
  {\fontsize{6.5}{7}\selectfont
    \begin{tabular}{|c|c|c|}
    \hline
    Network & Experiment 3 & Experiment 4 \\
    \hline
    Network 1 & 0.0988 & 0.0553 \\
    \hline
    Network 3 & 0.0908 & - \\
    \hline
    Network 4 & 0.0824 & 0.0529 \\
    \hline
    \end{tabular}%
    }
  \label{tab:var}%
  \vspace{-5mm}
\end{table}%

\section{Conclusion}

We showed that AU class imbalance and patterns are correlated, and the same patterns exist across multiple emotions, which can impact how AU detection methods learn. We validated this through multi- and single-AU experiments, showing performance decreases when networks are unable to learn the patterns. An interesting result from these experiments is the overall comparable decrease in accuracy of \textit{Network 2}, with single AU detection, compared to \textit{Network 1} (Fig. \ref{graph:balancedDataComparisons}). This has led us to a new hypothesis that specific network architectures are less important when AU patterns are detected compared to single AUs. Considering this, we will conduct evaluations on multiple network types for single AU and pattern detection.

To utilize the impact of AU patterns on detection, we proposed a new method that uses the AU patterns to directly train deep neural networks. We have shown that this method, compared to individual action units, can improve accuracy, as well as lower the overall variance among action units. In other words, the proposed approach helps to mitigate the negative impact that AU class imbalance has on detection results, as can be seen in Figs. \ref{graph:DistributionBP4D} and \ref{graph:DistributionDISFA}. This approach can be integrated into any deep neural network architecture helping to further improve state of the art results.

\bibliographystyle{ieeeTran}
\bibliography{references}
\end{document}